\pdfoutput=1
\documentclass[11pt]{article}

\usepackage[]{EMNLP2022}

\usepackage{times}
\usepackage{latexsym}

\usepackage[T1]{fontenc}
\usepackage[utf8]{inputenc}

\usepackage{microtype}

\usepackage{graphicx}
\usepackage{float}

\title{What is Wrong with Language Models that Can Not Tell a Story?}

\author{Ivan P. Yamshchikov \\
  Max Planck Institute for\\
 Mathematics in the Sciences\\
 Leipzig, Germany\\
 CEMAPRE, \\
 University of Lisbon, Portugal\\
  \texttt{ivan@yamshchikov.info} \\\And
   Alexey Tikhonov\\
Inworld.AI \\
Berlin, Germany \\
\texttt{altsoph@gmail.com}}

\begin{document}
\maketitle
\begin{abstract}
This paper argues that a deeper understanding of narrative and the successful generation of longer subjectively interesting texts is a vital bottleneck that hinders the progress in modern Natural Language Processing (NLP) and may even be in the whole field of Artificial Intelligence. We demonstrate that there are no adequate datasets, evaluation methods, and even operational concepts that could be used to start working on narrative processing.
\end{abstract}

\begin{figure*}
  \includegraphics[width=\textwidth]{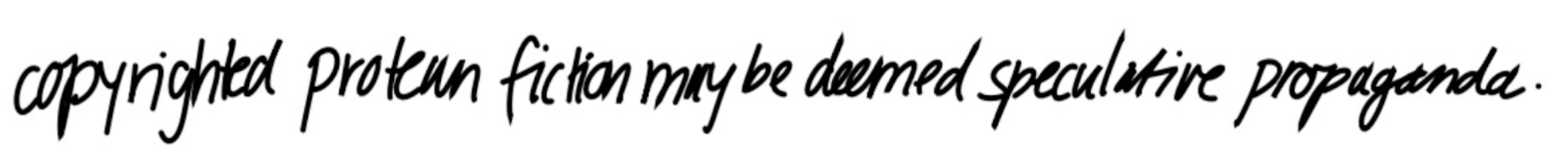}
  \caption{"Copyrighted protein fiction may be deemed speculative propaganda" — a line from a generative art project "Paranoid Transformer — a diary of an artificial neural network", \cite{agafonova2020paranoid}. The diary was generated end-to-end without any human post-processing and published as a hardcover book.}
  \label{fig:teaser}
\end{figure*}

\section{Introduction}
Perhaps since the linguistic turn in the early 20th century \cite{Wittgenstein1922}, human language is understood to be fundamental in framing human cognition. This epistemically distinguished position of language makes it tempting to equate intelligence with the capability to generate natural language. Famously Turing suggests in \cite{turing1950computing} that the ability to provide meaningful interaction in a natural language is crucial for artificial intelligence. Most modern researchers reduce the Turing test to the narrow frame of day-to-day conversations. Still, the original essay highlights that artificial intelligent agents should successfully imitate humans before any human judge. Turing explicitly mentions that such an agent should be able to perform creative tasks expressed in natural language. Framing the problem in Turing's original terms, one sees how far modern artificial systems are from fulfilling this vision. Modern chatbots can imitate human dialog in some contexts but have a hard time coming up with an exciting story \cite{van2019narrative} or a joke \cite{niculescu2021brief}. 

%The fundamental status of human language is solidified by recent neurobiological evidence that language is intertwined with human cognition \cite{lieberman2009foxp2}. 

Modern Natural Language Generation (NLG) is successfully exploiting significant increases in computational power and volume of training data \cite{brown2020language, raffel2020exploring, chowdhery2022palm, bajaj2022metro, zoph2022designing}; the field focuses on extensive computation-heavy solutions rather than on the statistical methods and mathematical models that could qualitatively change our understanding of language. In 1913 Andrey Markov invented his eponymous chains to analyze poetry. A century later, NLG, in a nutshell, exploits ideas that are not far from Markov's original intuition. The conceptual weakness of the current approach is qualitative and can not be healed by quantitative means. It is especially apparent in narrative processing, where attempts to automatically generate textual narratives either demand extensive human intervention, are employing domain-specific procedural generation, or use some predefined narrative structure as the backbone of the generated text \cite{van2019narrative}.

There are constant attempts to generate longer blocks of text\footnote{https://github.com/NaNoGenMo}, such as \citep{kedziorski2019understanding} or \citep{agafonova2020paranoid}. Yet, they succeed under certain stylistic and topical constraints that exclude the problem of narrative generation altogether. Although there are several recent results in the areas of suspense generation \citep{doust2017model}, narrative personalization \citep{wang2017interactive}, and generation of short context-based narratives \citep{womack2019interactive},  generating long stories is still a challenge \citep{van2019narrative}.

Though philosophers and linguists have tried to conceptualize the notions of plot, narrative arc, action, and actor for almost a century \cite{shklovsky1925theory,propp1968morphology,van1976philosophy}, few of these theoretical concepts could be instrumental for modern NLP. \cite{ostermann2019mcscript2} present a machine comprehension corpus for the end-to-end evaluation of script knowledge. The authors demonstrate that though the task is not challenging to humans, existing machine comprehension models fail to perform well. Despite these discouraging results, there are various attempts to advance narrative generation within the NLP community, see \cite{fan2019strategies,ammanabrolu2020story}. However, one still can hardly talk about narrative generation as a sub-field within NLP. It largely (and we believe, erroneously) remains a fringe research topic.

Yet we believe that the concept of {\em narrative} is crucial for further progress of NLP, and narrative generation should move to the pinnacle of the NLP community attention. This paper presents a set of questions that are vital for narrative processing to become a fully independent well-defined sub-field within NLP research. We start by presenting several arguments for why breakthroughs in narrative processing could be seminal for modern research of artificial intelligence in general. Then we move on to the nature of the bottlenecks that hinder the progress in narrative processing. We decompose the question "why we still do not have an algorithm to generate good stories?" into three systemic components: data, evaluation methods, and concepts. We argue that we face severe challenges on these three fronts, while only one of the areas (namely, data) is partially addressed. 

\section{On the Importance of Narrative}

Before addressing three fundamental bottlenecks that separate us from qualitatively new narrative generation models, let's briefly present a case on why narrative processing is pivotal for the further development of NLP. After all, recent years could be regarded as a triumph of language models driven by a distributional hypothesis \cite{harris1954distributional}. These models are profoundly local in terms of their input and training yet seem to be game-changers even out of the scope of classical NLP. For example, \cite{lu2021pretrained} demonstrate that pretraining on natural language can improve performance and compute efficiency on non-language downstream tasks. \cite{zeng2022socratic} propose a new approach to AI systems when new multimodal tasks are formulated as a guided language-based exchange between different pre-existing foundation models. \cite{tam2022semantic} discuss why and how language provides useful abstractions for exploration in reinforcement learning 3D environment. Do we really need narrative after all? May it be that all verbal cognition, similarly to politics, is local?

We advocate that narrative processing as a research field would have a seminal impact on two other core aspects of natural language processing that are badly needed to extend the adoption of NLP products and technologies. The first one is causality and natural language inference. Causal inference from natural language is vital for further progress in NLP, and to the moment, modern LMs are still inferior in this field. Though one would argue that narrative is but a sub-category within a broader family of causal texts, we argue that narrative generation is a perfect task to validate and test hypotheses around natural language inference paving the path to more explainable AI. The second field where narrative processing is essential is the further development of human-machine interaction. People are known to remember stories rather than facts \cite{wiig2012people}. NLP-based natural language interfaces demonstrate radically different properties: facts are far easier to process and "remember" than stories. These three bottlenecks make the narrative vital for further NLP progress.

%The second field where narrative processing might become pivotal is explainable AI. One could claim that one of the viable paths to explainable artificial intelligence goes through a set of dedicated models trained to communicate with humans in natural language, explaining particular aspects of a given decision. Such models would necessarily have to be capable of causal inference in natural language. 

 %Though technically, this leads to the same bottleneck that we have discussed above, we believe this field is so crucial for further development and adoption of artificial intelligence in the industry that it is worth an explicit mention here. Finally, 

\section{Where Do We Fail?}

This paper draws the research community's attention to fundamental cavities in our conceptual understanding, benchmarking, and evaluation in the area of narrative processing. We argue that these three significant layers need the urgent attention of the research community. This section discusses each of these layers in detail and suggests possible ways to move forward.

\subsection{Data}

Most available datasets designated as narrative datasets in the scientific literature are far from a common sense understanding of a "story." Some of the authors even call their datasets {\em scenarios} rather than stories of narratives. These datasets are also too small to be meaningfully used with the modern transformer-based language model. In \cite{regneri2010learning} authors collect  493 event sequence descriptions for 22 behavior scenarios. In \cite{modi2016inscript} author present InScript dataset that consists of 1,000 stories centered around 10 different scenarios.  \cite{wanzare2019detecting} provide 200 scenarios and attempt to identify all references to them in a collection of narrative texts. 

As we progress towards longer stories, the landscape of available data becomes far more deserted. \cite{mostafazadeh2016corpus} present a corpus of 50k five-sentence commonsense stories. MPST dataset that contains 14K movie plot synopses, \cite{kar2018mpst}, and WikiPlots\footnote{https://github.com/markriedl/WikiPlots} contains 112 936 story plots extracted from English language Wikipedia. Recently \cite{malysheva2021dyplodoc} provided a dataset of TV series along with an instrument for narrative arc analysis. These datasets are useful, yet a vast majority of the narrative datasets are only available in English. In \cite{tikhonov2021storydb} authors present StoryDB — a broad multilanguage dataset of narratives. With stories in 42 different languages, the authors try to amend the deficit of multilingual narrative datasets. 

Data is the only area of narrative processing where we can see some positive dynamics, yet one has to admit the following state of affairs. There is a limited number of datasets with longer narrative texts. The ones that are available are mostly in English and rarely include any human labeling regarding the narrative structure and/or their quality. There is minimal discussion of which narrative datasets the community needs to advance narrative generation.

\subsection{Evaluation}
Before we discuss narrative per se, let us first discuss evaluation techniques available for natural language generation in general. In \cite{hamalainen2021human} the authors review several recent generative papers. The methods are both automated and manual when native speakers are instructed to assess specific properties of the generated text. This review includes more than twenty papers on text generation evaluating several aspects of the generated texts with human labels. We believe that the scope of the paper represents the field at large. 

Let us look closely at the aspects of evaluation that are addressed in these 20+ papers on text generation. There is a variety of methods, approaches, and concepts here. For the details, we address the reader to \cite{hamalainen2021human} yet in the context of the current discussion; we take the liberty to categorize a vast majority of the proposed methods into five major groups.

\textbf{Fluency}; these are various ways to estimate if a generated text has grammatical and syntactic mistakes; these metrics are relatively well defined and could be automated to some extent; at least 13 papers out of the 23 NLG papers in the study are using one or several fluency metrics for evaluation.

\textbf{Topic/style/genre matching}; these metrics also could be automated and are usually based on some form of a pretrained classifier, see \cite{ficler2017controlling}; 12 papers in the study rely on one or several  evaluation criteria of this kind;

\textbf{Coherence}; this group of metrics is far more arbitrary. There seem to be at least three major types of coherence evaluation approaches. First, some form of coherence estimation on a level of linguistic pragmatics, namely, coherence of certain causal statements that include words like "hence/so/thus/etc." The second approach tries to estimate if the generated text is coherent with the general knowledge of the reader about the world. Naturally, these questions are far mode arbitrary, especially since fictional texts are based on the premise that they describe some alternative reality\footnote{Yet, we all have an intuitive understanding that some science fiction or fantasy novels are coherent, though non-realistic.}. Finally, the most abstract set of methods tries to ask the reader if the text is coherent within the internal logic of the "world" that it describes. This is the highest level of abstraction so far, and it naturally implies higher misalignment between human annotators and lower potential for automated evaluation. Even this brief overview demonstrates that there is no consensus on the coherence evaluation, yet 10 out of the 23 papers in the study used coherence evaluation.

\textbf{Overall emotional effect}; these metrics are harder to automate since they rely on human emotional response, yet provided enough human labels, one could train some classifier for such task;	11 out of the 23 papers in the study refer to some form of emotional effect evaluation.

\textbf{Novelty/originality/interestingness}; these metrics are even harder to formalize and use in an automated manner. Most of the papers imply certain novelty when they ask human labelers to assess the interestingness of the generated texts. However, human labelers could treat interestingness as a topic-related category. 7 out of 23 papers in the review used human evaluation of novelty.

The first two types of evaluation are the most dominant ones for automated evaluation methods, while coherence or novelty are rarely evaluated rigorously. Numerous NLG papers that use some automatic evaluation fit in these five categories. This list highlights that we hardly have any tools for assessing generated narratives. 

Coherence is something that a human can intuitively estimate. Yet, this evaluation is notoriously hard to automate. In the meantime, we still have a dubious understanding of the most straightforward tools, such as semantic similarity metrics for short texts, see \cite{yamshchikov2021style, solomon2021rethinking}. 

Novelty also depends on a deeper understanding of semantics, yet it might have an additional level of complexity. After all, human experience generally tells us that understanding something presented to you is, on average, a less challenging task than coming up with something new from scratch. 

Summing up, we have to conclude that out of all five groups of metrics used in the human evaluation, not a single one could both be fully automated and applied to narrative evaluation. They are all either automated but work on a lower level of shorter texts or deal with high-level conceptual questions yet are not quantified in a manner that allows automatic evaluation. This shocking observation leads us to the following logical conclusion: we can not explain to humans how one could evaluate a narrative. We lack a conceptual understanding of what it is and how to assess it.

\subsection{Concepts}

In a review paper \cite{gervas2019long} authors make a consistent case that "what people understand by the concept of storytelling is, in fact, a set of considerably diverse operations that are sometimes carried out in isolation to achieve simple stories or specific ingredients that might be a part of stories, and sometimes combined into the production of more elaborate stories."  In particular, the authors propose to "deconstruct" storytelling into the following set of approaches: stories as narrative Structures; stories as simulations; stories as evolving networks of character affinity; stories as narrations of observed facts; stories as suspense-driven entertainment.

Once again, a closer examination of the proposed taxonomy highlights similar problems that we saw in the evaluation. There are no uniformly agreed mechanisms for narrative representation that would have high coherence among human labelers. Most of the methods are either deeply subjective (see, for example, \cite{borjes1972} where the famous anthology of four plots was first presented) or are extremely low-level and work for causal inference on a short time scale yet could not be extended to the level of a short story, not no mention a novel.

We have to highlight here that each conceptual approach can show some practical results. Yet, there is no clear understanding of how these approaches structure a larger field of narrative processing that we argue should be the focus of NLP and AI communities in the nearest future. Is one of the approaches sufficient to obtain new models capable of generating an entertaining story? Do we need some combination of these pipelines described above? Should there be some qualitative and quantitative interaction between these pipelines? How should we organize it? There is not clear agreement on these basic questions in NLP community.

\section{Conclusion}
 This position paper makes two core statements:
\begin{itemize}
\item generation of novel entertaining narrative is a key task that might enable further progress of artificial intelligence across a number of fields and industries;
\item despite the vital importance of this task, modern NLP and AI communities are nowhere near to sharing a common understanding of which dataset could one use to work on narrative generation, how such tasks should be evaluated, and what are concepts the we have to define rigorously in order to work on these problems.
\end{itemize}

We hope this paper could spawn further discussion of these matters and might attract the NLP and AI community's attention to the issues around the narrative generation.

%\section*{Limitations}

%This is a position paper thus we do not see what the potential limitations could be. The only potential limitation might be the incompleteness of the list of relevant publications.

%\section*{Ethics Statement}
%This paper complies with the \href{https://www.aclweb.org/portal/content/acl-code-ethics}{ACL Ethics Policy}.

\bibliography{custom}
\bibliographystyle{acl_natbib}

%\appendix

%\section{Appendix}
%\label{appendix}

\end{document}